\documentclass[10pt,letterpaper]{article}
\PassOptionsToPackage{table}{xcolor}
\usepackage{times}
\usepackage[round,authoryear]{natbib}
\date{}
\usepackage[T1]{fontenc}
\usepackage{amsmath,amssymb}
\usepackage{graphicx}
\usepackage{wrapfig}
\usepackage{flafter}
\usepackage{tikz}
\newcommand{\bcirc}[1]{%
  \tikz[baseline=(char.base)]{
    \node[shape=circle, fill=black, inner sep=1.2pt] (char)
    {\textcolor{white}{\scriptsize #1}};
  }%
}
\usepackage{xcolor}
\usepackage{array,booktabs,tabularx,etoolbox}
\AtBeginEnvironment{tabularx}{\renewcommand{\arraystretch}{1.18}}
\AtBeginEnvironment{tabular*}{\renewcommand{\arraystretch}{1.14}\setlength{\tabcolsep}{4pt}}
\AtBeginEnvironment{table}{\setlength{\belowcaptionskip}{5pt}}
\usepackage{hyperref}
\usepackage{url}
\definecolor{paperblue}{HTML}{397B94}
\definecolor{papercoral}{HTML}{B95450}
\definecolor{papergray}{HTML}{F0F0F0}
\hypersetup{colorlinks=true,citecolor=blue,linkcolor=paperblue,urlcolor=paperblue}
\newcommand{\rmuse}{r_{\mathrm{M}}}
\newcommand{\rmyield}{r_{\mathrm{sup}}}
\newcommand{\rmabsent}{r_{\mathrm{obl}}}
\newcommand{\rmoff}{r_{\mathrm{irr}}}

\title{\raisebox{-0.12em}[0pt][0pt]{\includegraphics[height=1.05em]{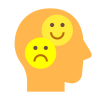}}\hspace{0.3em}PairPref: When Should Memory Guide the Answer? A Benchmark for Contextual Preference Use}
\author{Mingfei Lu \qquad Mengjia Wu \qquad Yi Zhang\\[0.5em]
\normalsize AAII, University of Technology Sydney\\[0.3em]
\small \texttt{mingfei.lu@student.uts.edu.au}\\
\small \texttt{mengjia.wu@uts.edu.au} \quad \texttt{yi.zhang@uts.edu.au}}

\begin{document}
\maketitle
\begin{abstract}
Memory-augmented assistants use retrieved preferences to guide their responses. Task circumstances, such as who a recommendation is for or what resources are available, can make a remembered preference inappropriate to apply. Applying it anyway can produce an answer that conflicts with the current task’s needs. Memory benchmarks typically test whether systems store and retrieve preferences, with less attention to when those preferences should guide an answer. We introduce PairPref, a benchmark of contextual preference use. Each pair holds the stored preference, request wording, and four candidate replies fixed while varying the task circumstances. The preference remains valid but should guide the answer in only one of the two situations. In the selection track, models must choose the reply that applies the preference only where appropriate. In the free-generation track, they must decide when to apply it without seeing candidate replies. Both tracks use the same 1,227 pairs across 45 preferences and eight situation categories. We evaluate eight models, most of which achieve selection scores ($\Delta$) of 51 to 65 points. In free generation, however, both responses are appropriate for their respective situations in only 3.6\% to 18.3\% of pairs retained after hedging exclusions. Models continue to apply the preference in both situations even with fewer retrieved memories, alternative presentation formats, and a stricter prompt. Models still struggle to judge when preferences apply and act accordingly.
\end{abstract}
\section{Introduction}
\begin{figure}[ht]
\centering
\includegraphics[width=\linewidth]{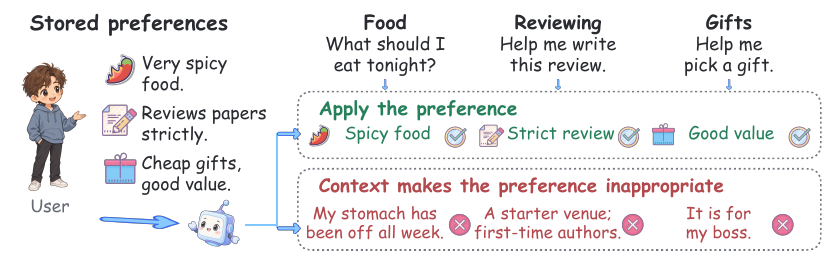}
\caption{User preferences remain valid across situations, but their applicability changes with context. The same request may call for applying a preference in one situation and setting it aside in another. PairPref evaluates whether assistants make this distinction across paired situations.}
\label{fig:example}
\end{figure}

Language-model assistants increasingly support users in everyday tasks, from choosing meals to reviewing papers and selecting gifts. Across repeated interactions, they accumulate user preferences that help personalize their responses \citep{zhong2024memorybank,xu2025amem,lu2026fluxmem}. These preferences can remain valid even when the situation calls for a different response. An assistant must therefore judge when a remembered preference should guide its answer (Figure~\ref{fig:example}).

Memory and personalization benchmarks fall into three lines. (1) \emph{Long-term memory benchmarks} evaluate recall and reasoning over extended interactions, as in LoCoMo and LongMemEval \citep{maharana2024locomo,wu2025longmemeval}. (2) \emph{Preference-following benchmarks} assess preference inference, tracking, and application, including PrefEval, PersonaMem, and CUPID \citep{zhao2025prefeval,jiang2025personamem,kim2025cupid}. (3) \emph{Memory-use boundary benchmarks} examine inappropriate personalization and disclosure \citep{hu2026opbench,mireshghallah2025cimemories,pulipaka2026persistbench,xu2026hush}. Within this line, RPEval evaluates whether a preference should be ignored, support a response, or dominate it \citep{feng2026rpeval}, while BenchPreS tests expression preferences under formal communication norms \citep{yoon2026benchpres}. We test a contextual contrast (Table~\ref{tab:benchmark_protocols}):
\begin{center}
\textbf{When the preference and request remain unchanged, can an assistant apply the preference in one situation and set it aside in another, responding appropriately in both cases?}
\end{center}

To address this question, we introduce PairPref, a benchmark of contextual preference use comprising 1,227 pairs across 45 preferences and eight situation categories. Each pair holds a still-valid preference and a request fixed while varying the situation so that the preference should guide the answer in one case but not the other. The selection track presents the same four candidate replies in both situations, testing whether the model changes its choice appropriately. The free-generation track removes the candidates and tests whether the model produces appropriate answers for both situations. Together, the two tracks assess whether models adjust preference use to context, both when selecting a reply and when composing their own.

We evaluate eight models and find that contextual preference use remains challenging. Seven models achieve selection scores ($\Delta$) between 51.3 and 64.9 points, yet both generated responses are appropriate for their respective situations in only 3.6\%--18.3\% of pairs across the eight models. Models apply the preference in both situations in 71.0\%--92.0\% of generated pairs. This pattern persists with fewer retrieved memories, alternative presentation formats, and a stricter prompt. These results highlight the difficulty of translating situational differences into appropriate preference use.

Our contributions are summarized as follows:

\noindent\bcirc{1} \textbf{Contextual preference benchmark.}
We introduce PairPref to test contextual applicability while keeping a still-valid preference and the request fixed.

\noindent\bcirc{2} \textbf{Paired evaluation protocol.}
We combine reply selection and free generation on the same situation pairs to measure whether both responses are appropriate.

\noindent\bcirc{3} \textbf{Empirical findings.}
We evaluate eight models and identify persistent preference application across situations, even when the context calls for setting the preference aside. This behavior persists across memory-pool sizes, presentation formats, and prompting conditions.

\section{Related work}
\label{sec:related}
\begin{table}[!t]
\caption{Capabilities assessed by preference and memory benchmarks.}
\label{tab:benchmark_protocols}
\centering
\small
\setlength{\tabcolsep}{3pt}
\renewcommand{\arraystretch}{1.2}
\newcommand{\capyes}{\raisebox{-0.2ex}{\includegraphics[height=1.45ex]{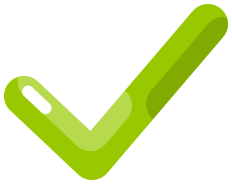}}}
\newcommand{\cappart}{\tikz[baseline=-0.5ex]{\fill[papergray] (0,0) circle (0.65ex);\fill[paperblue] (0,0.65ex) arc (90:270:0.65ex) -- cycle;\draw[paperblue,line width=0.3pt] (0,0) circle (0.65ex);}}
\newcommand{\capnone}{\textemdash}
\begin{tabularx}{\linewidth}{@{}>{\raggedright\arraybackslash}p{1.95cm}>{\raggedright\arraybackslash}p{3.8cm}*{4}{>{\centering\arraybackslash}X}@{}}
\toprule
Benchmark & Main evaluation target & Preference following & Preference suppression & Selection & Generation \\
\midrule
\multicolumn{6}{@{}l}{\emph{Long-term memory}} \\
LoCoMo & Conversational recall & \capnone & \capnone & \capnone & \capnone \\
LongMemEval & Long-term memory & \cappart & \capnone & \capnone & \capnone \\
\midrule
\multicolumn{6}{@{}l}{\emph{Preference following}} \\
PrefEval & Preference following & \capyes & \capnone & \capnone & \capnone \\
PersonaMem & Evolving user profiles & \capyes & \cappart & \cappart & \cappart \\
CUPID & Contextual preferences & \capyes & \cappart & \capnone & \cappart \\
\midrule
\multicolumn{6}{@{}l}{\emph{Memory-use boundaries}} \\
RPEval & Preference-use strategies & \capyes & \capyes & \cappart & \cappart \\
OP-Bench & Over-personalization & \capnone & \capyes & \capnone & \cappart \\
BenchPreS & Communication norms & \capyes & \capyes & \capnone & \cappart \\
CIMemories & Contextual disclosure & \capnone & \capnone & \capnone & \capnone \\
PersistBench & Leakage and sycophancy & \cappart & \cappart & \capnone & \cappart \\
HUSH-Bench & Sensitive-history boundaries & \cappart & \cappart & \capnone & \cappart \\
\midrule
\rowcolor{papergray}
\textbf{PairPref} & Preference applicability & \capyes & \capyes & \capyes & \capyes \\
\bottomrule
\end{tabularx}
\par\smallskip
\begin{minipage}{\linewidth}
\footnotesize
\capyes\ Direct evaluation; \cappart\ related coverage; \capnone\ outside the evaluation target. Selection and generation assess when to apply or withhold the same still-valid preference across situations with a fixed request. Related coverage uses different controls or task definitions; see Appendix~\ref{app:benchmark_distinctions}.
\end{minipage}
\par\vspace{-12pt}
\end{table}

Preference and memory benchmarks increasingly assess appropriate use as well as recall. Table~\ref{tab:benchmark_protocols} compares their coverage of preference following, suppression, selection, and generation. Appendix~\ref{app:benchmark_comparison} gives evaluation protocols, response formats, dataset sizes, and validation procedures.

\paragraph{Remembering and applying user information.}
Multi-session dialogue tests continuity across chats \citep{xu2022msc}. LoCoMo and LongMemEval evaluate long-term conversational memory; the latter also tests preference questions and abstention when information is unavailable \citep{maharana2024locomo,wu2025longmemeval}. HaluMem localizes failures to extraction, update, and question answering \citep{chen2025halumem}. LaMP and LongLaMP test profile-conditioned classification and long-form generation \citep{salemi2024lamp,kumar2024longlamp}. PrefEval evaluates explicit and implicit preference following through generation and multiple-choice tasks, including response helpfulness \citep{zhao2025prefeval}. PersonaMem tracks evolving user profiles, while CUPID tests inference and application of context-dependent preferences \citep{jiang2025personamem,kim2025cupid}. PairPref holds a preference's validity fixed and tests how its applicability changes with the situation.

\paragraph{Deciding when and how to use memory.}
Benchmarks assess both preference strategies and information-use boundaries. For preference strategies, RPEval evaluates whether preferences should be ignored, support an answer, or dominate it \citep{feng2026rpeval}; BenchPreS measures appropriate application and suppression under formal communication norms \citep{yoon2026benchpres}. For over-personalization and information disclosure, OP-Bench tests irrelevance, sycophancy, and repetition \citep{hu2026opbench}; PersistBench tests leakage and sycophancy alongside beneficial memory use \citep{pulipaka2026persistbench}. CIMemories measures inappropriate disclosure and necessary information coverage \citep{mireshghallah2025cimemories}, while HUSH-Bench tracks sensitive-history use under different memory-access and invitation conditions \citep{xu2026hush}. PairPref isolates a situational reversal within these contextual boundaries: the same still-valid preference and request warrant application in one situation and non-application in another. It evaluates both situations jointly, using fixed candidate replies for selection and the same pairs for free generation. See Appendix~\ref{app:benchmark_distinctions} for details.

\paragraph{Organizing and retrieving memories.}
Memory systems study how records are formed, stored, and retrieved \citep{hu2025memorysurvey}. Generative agents store experiences and reflect on them \citep{park2023generative}. Tiered stores separate short- and long-term records \citep{packer2023memgpt,kang2025memoryos}; linked notes and temporal graphs preserve relations among entries \citep{xu2025amem,rasmussen2025zep}. Retrieval-augmented generation supplies external text \citep{lewis2020rag}, while HippoRAG and its successor use structured retrieval as non-parametric memory \citep{gutierrez2024hipporag,gutierrez2025hipporag2}. Other systems reduce memory-processing costs \citep{chhikara2025mem0,fang2026lightmem} or learn what to write, retrieve, and fuse \citep{yan2026memoryr1,du2025memr3,lu2026fluxmem,zhang2026memgate}. FluxMem selects memory structures according to interaction features \citep{lu2026fluxmem}, while SEER combines visual-text compression with selective retrieval of query-relevant text \citep{xu2026seer}. PairPref tests whether models adjust preference use to the situation given fixed memories.

\paragraph{Structured reasoning and evidence use.}
Related questions about extracting and selecting useful information arise in domain-specific systems. From Query to Counsel organizes legal facts and intents into an element graph for multi-agent consultation \citep{lu2026query}; bias-aware citation prediction combines multi-agent feature extraction with graph representations \citep{lu2026newborn}. In graph fraud detection, diffusion-guided learning addresses sparse supervision through feature augmentation \citep{liu2026beyond}, whereas PREF-Gate constrains label-derived evidence by provenance and selects its use through validation \citep{liu2026pref}. These approaches concern task representations and evidence quality. PairPref examines the applicability of a supplied user preference: even correctly retrieved information may be unsuitable for the current response.

\section{Benchmark construction}
\label{sec:design}
PairPref tests whether assistants adjust preference use when the situation changes but the preference remains valid. Each pair fixes a preference and a request, then contrasts a situation where the preference should guide the answer with one where it should be set aside. Selection presents the same four candidate replies in both situations; free generation removes the candidates. Figure~\ref{fig:workflow} summarizes construction, validation, and both evaluation tracks.

\subsection{Task formulation}
Let $M_i$ denote a persistent preference, $q_i$ a shared request, and $C_i^+$ and $C_i^-$ situations that warrant applying and setting aside the preference, respectively. Each pair is
\begin{equation}
 x_i=(M_i,q_i,C_i^+,C_i^-,\mathcal{R}_i),
 \qquad \mathcal{R}_i=\{\rmuse,\rmyield,\rmabsent,\rmoff\}.
\end{equation}
The four replies distinguish preference use from acknowledgment and task completion. The $\rmuse$ reply fully follows the preference; $\rmyield$ departs from it and acknowledges doing so; $\rmabsent$ departs without acknowledgment; and $\rmoff$ answers an unrelated task. The two departing replies may recommend the same action. This contrast separates acknowledgment from the action itself, while the off-task reply distinguishes withholding a preference from completing the request.

Selection tests whether the model changes its reply appropriately across the pair. Free generation tests whether it produces appropriate responses without candidates. Both tracks assess success on the two situations jointly. The preference, request, shared task facts, and companion memories remain fixed, isolating the effect of the situation on preference use.

\subsection{Paired instance construction}

\begin{figure}[t]
\centering
\includegraphics[width=\linewidth]{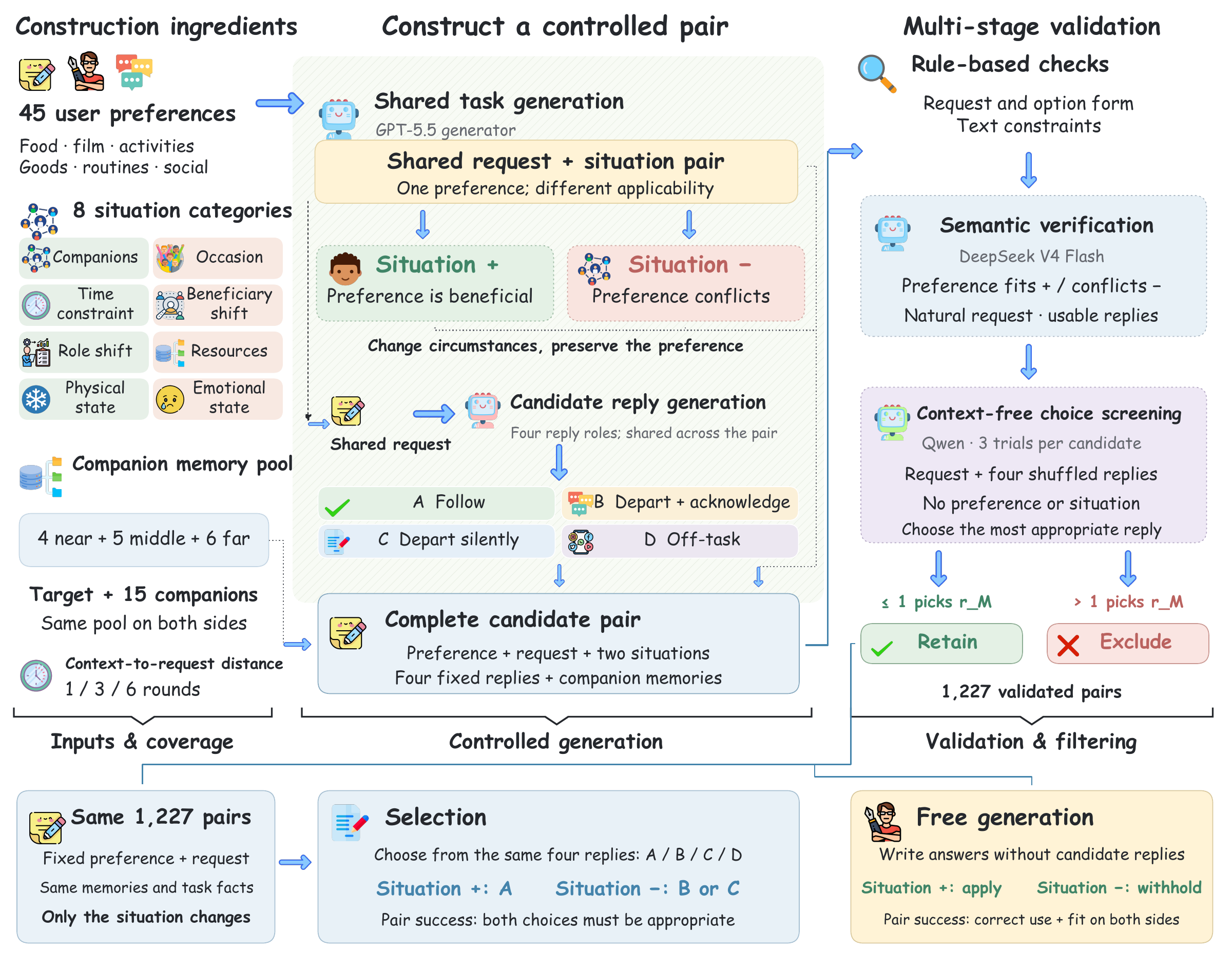}
\caption{PairPref construction and evaluation. Controlled pairs pass three validation stages and support selection and free generation on the same situations.}\label{fig:workflow}
\label{fig:data_construction}
\end{figure}
\paragraph{Generating situations and replies.}
We construct pairs around 45 preferences across six domains and eight situation categories. GPT-5.5 generates a shared request and two situations in which the same preference differs in applicability, then produces four candidate replies representing preference use, acknowledged departure, unacknowledged departure, and an unrelated task. The request and replies remain fixed across the pair. Situations convey circumstances that affect the decision rather than directly instructing the assistant to follow or ignore the preference.

\paragraph{Supplying companion memories.}
Each target preference is accompanied by four near, five middle, and six far memories. The resulting 16-memory pool contains the same entries on both sides and is supplied directly to the model. The target preference is therefore available in both situations; the task is to judge whether to apply it. Context-to-request distances vary across one, three, and six rounds, with filler exchanges separating context from the request.

\subsection{Multi-stage validation}
\label{sec:quality}
We validate candidate pairs through three complementary stages: rule-based checks, model-based semantic verification, and context-free choice screening (Figure~\ref{fig:data_construction}).

\paragraph{Rule-based checks.}
Programmatic checks screen candidate text and option form, including reply lengths, repeated wording, and explicit cues that reveal the intended preference decision. These checks filter surface violations before semantic verification.

\paragraph{Semantic verification.}
DeepSeek V4 Flash checks whether the situations support the intended change in applicability, whether the shared request is natural in both situations, and whether the replies meet their intended roles. The checks also assess whether the constraint requires reasonable inference, whether preference use remains feasible on the positive side, and whether both sides have usable answers. Appendix~\ref{app:construction_rubric} details the checks and model configurations.

\paragraph{Context-free choice screening.}
To screen for candidate cues that bypass the contextual judgment, Qwen receives only the request and four replies, without the preference or either situation, and selects the most appropriate reply. Each candidate is tested three times and excluded if the preference-enacting reply is selected more than once. The 1,227 retained pairs pass all construction filters. Separately, two annotators independently assessed 200 sampled pairs after situation revisions, reaching 99.25\% applicability agreement ($\kappa=0.985$). Appendix~\ref{app:human_validation} reports this revised-sample assessment separately from the collection used for model evaluation.

\subsection{Dataset coverage}
PairPref contains 1,227 pairs covering 45 preferences across food, film, activities, goods, routines, and social interaction. Eight situation categories vary companions, occasion, time constraints, beneficiary, role, resources, physical state, and emotional state. Positive situations comprise 620 clean and 607 distractor cases: the latter introduce an additional situational cue while retaining the preference's applicability. Both evaluation tracks use the same pairs. Appendix~\ref{app:coverage} reports the category and domain distributions.

\section{Experiments}
\label{sec:legacy}
\subsection{Experimental setup}
\label{sec:pending}
\paragraph{Models and inputs.}
We evaluate eight models on both tracks of PairPref: Claude Opus 5 \citep{anthropic2026opus5}, Kimi K3 \citep{moonshot2026kimik3}, GPT-6 Astra \citep{openai2026astra}, GLM-5.3 \citep{zai2026glm53}, Gemini 3.5 Flash \citep{google2026gemini35}, DeepSeek V4 Pro \citep{deepseek2026v4}, Qwen 3.8 Max \citep{qwen2026qwen38}, and Grok 4.6 \citep{xai2026grok46}. Each input contains the target preference and 15 companion memories as a system list, followed by a situation, intervening dialogue turns, and the shared request. The default instruction asks models to use memories when relevant without forcing them into the answer. The two situations are evaluated in separate calls, with intended reply roles and side labels hidden. Selection provides the four candidate replies, whereas free generation omits them. See Table~\ref{tab:models} for the models and Appendix~\ref{app:fullreg-settings} for implementation details.

\paragraph{Metrics.}
The positive situation calls for using the preference; the negative situation calls for setting it aside. Let $u^+$ and $u^-$ denote preference use on these two sides. Selection assigns $u=1$ when the model chooses $\rmuse$ and 0 otherwise. Generation assigns $u\in\{0,0.5,1\}$ for no reliance, implicit reliance, and explicit or central reliance on the preference, respectively.

We report three measures of preference use. \emph{Use$^+$} ($\mathrm{MU}^+=\mathbb{E}[u^+]$) measures whether models use the preference when it fits; higher is better. \emph{Use$^-$} ($\mathrm{MI}^-=\mathbb{E}[u^-]$) measures whether they continue using it when it should be set aside; lower is better. \emph{Selectivity} ($\Delta=\mathrm{MU}^+-\mathrm{MI}^-$) measures the separation between these behaviors. Always using or never using the preference both yield $\Delta=0$; perfect separation yields 100 points. In selection, these averages are selection rates; in generation, they average graded uptake. Selection $\Delta$ counts off-task replies as non-use, not errors.

Selection \emph{Pair success} is the percentage of pairs with $\rmuse$ selected in the positive situation and either $\rmyield$ or $\rmabsent$ in the negative situation. Selecting the off-task reply never counts as success.

For free generation, we evaluate the two answers in each pair jointly. \emph{Use both} (Always-use) is the percentage of pairs in which the model uses the preference in both situations. Since the negative situation calls for setting it aside, this measures persistent preference use despite the contextual change; lower is better. \emph{Pair success} (Gen-OK) is the percentage of pairs satisfying all three conditions: the positive answer uses the preference, the negative answer does not, and both answers complete the request under their respective situations. A failure on either side makes the pair unsuccessful; higher is better. These metrics are not complements: avoiding the preference in both answers lowers Use both but does not constitute Pair success, nor does withholding it in an off-task answer.

Formally, Use both requires $u^+\ge0.5$ and $u^-\ge0.5$. Pair success requires $u^+\ge0.5$, $u^-=0$, and a situational-fit score of 1 on both sides. Each rate uses retained pairs with all required judgments.

Selection is scored directly from the chosen reply, while generated answers are evaluated by DeepSeek V4 Flash using separate judgments of preference use and situational fit.

\paragraph{Included pairs and uncertainty.}
We compute paired results only when both sides have valid outputs. We further exclude a generated pair if either answer offers both a preference-based and a generic alternative without committing to one, such as ``Choose X if alone, but Y if with company.'' This hedging rule distinguishes a contextual decision from covering both possibilities. Table~\ref{tab:models} reports the remaining number of pairs for each model; generation rates are calculated over these retained pairs, with Pair success also requiring both fit judgments. For example, a Pair success of 10\% means that one in ten scored pairs satisfies all success conditions. The 95\% confidence intervals use 1,000 bootstrap resamples, clustering pairs by target preference to account for shared content.

\subsection{Models struggle with contextual preference use in both tracks}
\suppressfloats[t]
\begin{table}[t]
\caption{Main results on PairPref.}
\label{tab:models}
\centering\small
\newcommand{\selscore}[2]{\shortstack{#1\\[-1pt]{\scriptsize\textcolor{black!60}{[#2]}}}}
\begin{tabular*}{\linewidth}{@{\extracolsep{\fill}}lrrc >{\columncolor{paperblue!8}}r rr>{\columncolor{paperblue!8}}r@{}}
\toprule
& \multicolumn{4}{c}{\textbf{Selection}} & \multicolumn{3}{c}{\textbf{Free generation}} \\
\cmidrule(lr){2-5}\cmidrule(lr){6-8}
Model & Use$^+$ $\uparrow$ & Use$^-$ $\downarrow$ & Selectivity $\uparrow$ & \shortstack{\textbf{Pair}\\\textbf{success} $\uparrow$} & \shortstack{Retained\\pairs} & \shortstack{Use both\\$\downarrow$} & \shortstack{\textbf{Pair}\\\textbf{success} $\uparrow$} \\
\midrule
Claude Opus 5 & 76.4 & \textbf{11.5} & \selscore{\textbf{64.9}}{58.2--70.6} & \textbf{66.4} & 1,067 & 88.7 & 9.3 \\
Kimi K3 & 76.0 & 18.3 & \selscore{57.6}{50.8--63.6} & 59.5 & 1,091 & 83.6 & 10.6 \\
GPT-6 Astra & 71.2 & 14.7 & \selscore{56.6}{49.8--62.6} & 57.9 & 1,068 & \textbf{71.0} & 17.9 \\
GLM-5.3 & 75.1 & 18.4 & \selscore{56.6}{50.3--61.9} & 58.4 & 1,056 & 90.8 & 5.7 \\
Gemini 3.5 Flash & \textbf{78.9} & 22.9 & \selscore{56.0}{49.1--62.5} & 57.9 & 1,065 & 75.1 & \textbf{18.3} \\
DeepSeek V4 Pro & 76.4 & 23.0 & \selscore{53.5}{47.4--58.6} & 55.2 & 1,107 & 81.3 & 13.5 \\
Qwen 3.8 Max & 71.6 & 20.3 & \selscore{51.3}{45.0--56.9} & 54.1 & 1,011 & 81.2 & 12.7 \\
Grok 4.6 & 75.1 & 57.4 & \selscore{17.7}{12.2--22.3} & 23.6 & 1,153 & 92.0 & 3.6 \\

\bottomrule
\end{tabular*}
\par\smallskip
\begin{minipage}{\linewidth}\footnotesize
Rates are percentages; selectivity ($\Delta$) is in percentage points, with 95\% CIs below. Generation rates use retained pairs. Bold marks the best value per metric. Metric definitions, scoring, and exclusions are in Section~\ref{sec:pending}.
\end{minipage}
\end{table}

\begin{figure}[t]
\centering
\includegraphics[width=\linewidth]{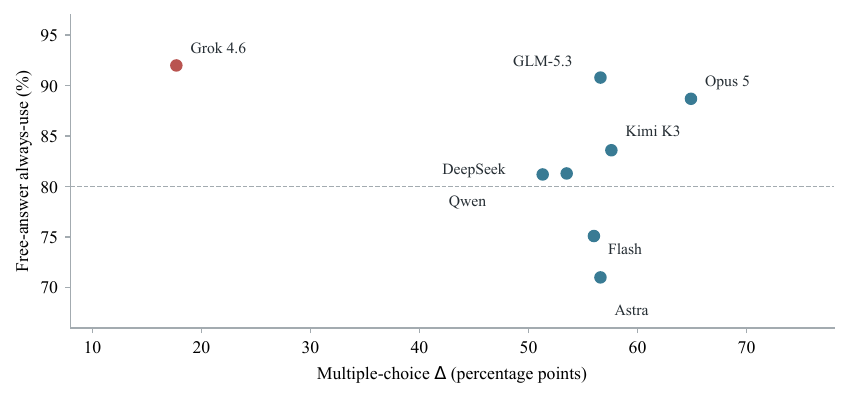}
\caption{Multiple-choice $\Delta$ versus free-answer always-use on the same pairs (Table~\ref{tab:models}). Contextual separation in selection coexists with frequent preference use in both generated answers.}
\label{fig:tracks}
\end{figure}
Contextual preference use remains challenging in both tracks (Table~\ref{tab:models}). Selection Pair success ranges from 23.6\% to 66.4\%, while generation Pair success is 3.6\%--18.3\%. Even the strongest selection result leaves roughly one third of pairs unsuccessful. These rates follow each track's scoring and inclusion rules and are not directly comparable across tracks.

The predominant generation pattern is to use the preference in both situations: Use both reaches 71.0\%--92.0\% across the eight models on retained generation pairs (Figure~\ref{fig:tracks}). Even when we restrict the analysis to pairs on which the same model succeeds on both selection sides, this proportion remains 71.8\%--92.2\%. Selecting the intended replies does not ensure appropriate preference use in answers generated for the same situations. See Appendix~\ref{app:cross-track} for matched counts and model results.

\paragraph{An illustrative case.}
A user who drinks only hand-poured single-origin coffee asks what to order at a market counter. With sufficient credit, Claude Opus 5 selects the pour-over reply; when the wristband has only enough credit for the self-serve urn, it selects the house-drip reply. Its generated positive answer also recommends pour-over. In the negative situation, however, it begins ``Skip the urn,'' rejects the available coffee because it does not match the user's preference, and recommends postponing coffee to visit another shop. The answer preserves the preference by moving the purchase elsewhere rather than choosing within the stated resource constraint. The judge records preference use on both sides and failed situational fit on the negative side (full example in Appendix~\ref{app:coffee-example}).

\par\ifdim\dimexpr\pagegoal-\pagetotal\relax<30\baselineskip\newpage\fi
\subsection{Effects of situation and memory}
\begin{table}[!ht]
\caption{Situation and memory removal.}
\label{tab:validity}
\centering\footnotesize
\begin{tabular*}{\linewidth}{@{\extracolsep{\fill}}lr>{\columncolor{paperblue!8}}r@{\hspace{14pt}}lrr>{\columncolor{paperblue!8}}r>{\columncolor{paperblue!8}}r@{}}
\cmidrule{1-3}\cmidrule{4-8}
\multicolumn{3}{@{}l}{\textbf{A. Selection: selectivity}} & \multicolumn{5}{l}{\textbf{B. Generation: preference uptake}} \\
\cmidrule{1-3}\cmidrule{4-8}
& Full input & No situation & & \multicolumn{2}{c}{Full input} & \multicolumn{2}{c}{No memory} \\
\cmidrule(lr){2-2}\cmidrule(lr){3-3}\cmidrule(lr){5-6}\cmidrule(lr){7-8}
Model & $\Delta$ & $\Delta$ & Model & Use$^+$ & Use$^-$ & Use$^+$ & Use$^-$ \\
\cmidrule{1-3}\cmidrule{4-8}
Gemini 3.5 & 56.0 & 0.0 & Gemini 3.5 & 78.8 & 64.7 & 11.1 & 7.5 \\
Kimi K3 & 57.6 & -0.2 & Kimi K3 & 84.4 & 77.6 & 15.1 & 13.1 \\
DeepSeek V4 & 53.5 & -0.5 & DeepSeek V4 & 84.1 & 73.4 & 11.7 & 9.3 \\
GLM-5.3 & 56.6 & -0.7 & GLM-5.3 & 87.5 & 84.2 & 16.3 & 15.8 \\
GPT-6 Astra & 56.6 & 1.4 & GPT-6 Astra & 78.3 & 67.4 & 11.4 & 11.3 \\
Qwen 3.8 & 51.3 & --- & Qwen 3.8 & 83.9 & 75.6 & 11.7 & 11.4 \\
\cmidrule{1-3}\cmidrule{4-8}
\end{tabular*}
\par\smallskip
\begin{minipage}{\linewidth}\footnotesize
A: $\Delta$ in percentage points; --- means not evaluated. B: mean preference uptake (0--100). No-memory uptake measures alignment with the withheld preference. Blue marks removals. Details: Appendix~\ref{app:removal-details}.
\end{minipage}
\end{table}

Removing the situation reduces selection selectivity to near zero; removing the memory block markedly reduces alignment between generated answers and the target preference (Table~\ref{tab:validity}). These controls show that contextual separation in selection depends on the situation, while frequent preference-related content in generation is associated with the supplied memory.

\subsection{Pool size does not explain preference misuse}
\suppressfloats[t]
A shorter pool preserves negative-side selection of the preference. Position in a long context can change what a model uses \citep{liu2024lostmiddle,hsieh2024ruler}. Top-5 retains the target preference and four near companions. Negative-side use increases on four of the five evaluated models; GPT-6 Astra decreases from 14.7\% to 11.7\% (Table~\ref{tab:pool}). Pair success changes by $-2.0$ to $+4.6$ percentage points across these models, with no consistent improvement (selectivity is reported in Appendix~\ref{app:pool-selectivity}). Replacing companions with 15 far items leaves $\Delta$ in the same range (Appendix~\ref{app:fullreg-far}). Misuse persists from a 16-item list to five related items.

\suppressfloats[t]
\begin{figure}[t]
\centering
\includegraphics[width=0.90\linewidth]{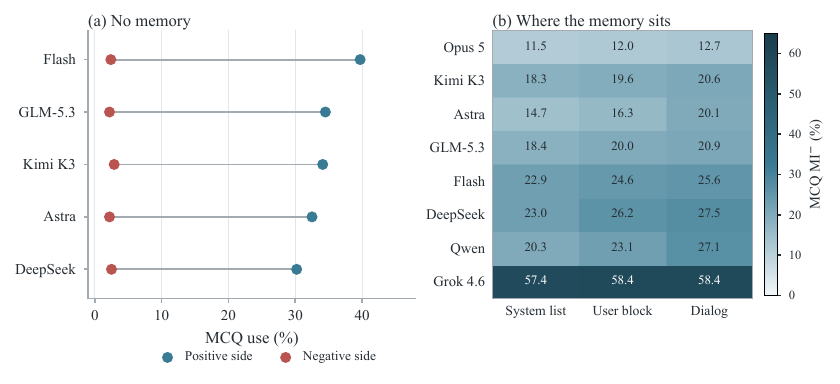}
\caption{Two controls on multiple choice. (a) Removing memory preserves positive-side but reduces negative-side selection. (b) Negative-side use remains similar across memory presentations.}
\label{fig:controls}
\vspace{-10pt}
\end{figure}
\begin{table}[t]
\caption{Effect of memory-pool size on selection performance.}
\label{tab:pool}
\centering\small
\begin{tabular*}{\linewidth}{@{\extracolsep{\fill}}lrr>{\columncolor{paperblue!8}}rrr>{\columncolor{paperblue!8}}r@{}}
\toprule
& \multicolumn{3}{c}{\textbf{Full pool (16)}} & \multicolumn{3}{c}{\textbf{Smaller pool (5)}} \\
\cmidrule(lr){2-4}\cmidrule(lr){5-7}
Model & Use$^+$ $\uparrow$ & Use$^-$ $\downarrow$ & \shortstack{\textbf{Pair success}\\$\uparrow$} & Use$^+$ $\uparrow$ & Use$^-$ $\downarrow$ & \shortstack{\textbf{Pair success}\\$\uparrow$} \\
\midrule
Gemini 3.5 Flash & 78.9 & 22.9 & 57.9 & 81.5 & 25.8 & 57.8 \\
DeepSeek V4 Pro & 76.4 & 23.0 & 55.2 & 78.6 & 27.8 & 53.1 \\
Kimi K3 & 76.0 & 18.3 & 59.5 & 79.7 & 20.7 & 61.3 \\
GLM-5.3 & 75.1 & 18.4 & 58.4 & 77.7 & 21.0 & 59.0 \\
GPT-6 Astra & 71.2 & 14.7 & 57.9 & 73.3 & 11.7 & 62.5 \\
\bottomrule
\end{tabular*}
\par\smallskip
\begin{minipage}{\linewidth}\footnotesize
Rates in \%; $n=1{,}227$, except GLM-5.3 with the smaller pool ($n=1{,}226$; one incomplete output).
\end{minipage}
\par\vspace{-18pt}
\end{table}

\subsection{Stricter prompting changes selection more than generation}
\begin{wrapfigure}{r}{0.5\textwidth}
\vspace{-8pt}
\centering
\includegraphics[width=\linewidth]{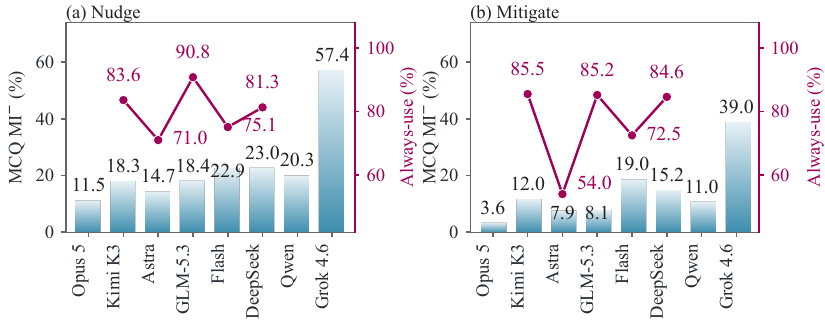}
\vspace{-1em}
\caption{Prompt wording. Bars are negative-side multiple choice (left axis). The line shows free-answer always-use (right).}
\label{fig:prompt}
\vspace{-1em}
\end{wrapfigure}
The default relevance instruction does not account for the selection pattern: removing the ``use when relevant'' clause changes multiple-choice negative use by at most a few points (Appendix~\ref{app:fullreg-extra}). Explicitly distinguishing truth from applicability in the mitigate instruction lowers multiple-choice misuse---Opus 5 from 11.5\% to 3.6\%, GLM-5.3 from 18.4\% to 8.1\%---and also lowers positive-side use for seven of eight models. Flash is the exception, moving from 78.9\% to 80.3\%. On free answers, always-use is 54.0\%--85.5\% for the five completed models (Figure~\ref{fig:prompt}). Astra moves from 71.0\% to 54.0\% always-use and from 17.9\% to 26.8\% Gen-OK. A stricter withhold prompt changes letter selection more than written recommendations.

\subsection{Preference misuse persists across memory presentations}
Multiple-choice misuse persists beyond the system-list format. Moving the same 16 memories into a user block or prior dialogue turns leaves negative-side use in the same band (Figure~\ref{fig:controls}b). Several models increase slightly (Qwen 3.8 Max 20.3\% to 27.1\% under dialog). These presentations rule out dependence on the system-list format.

\subsection{Topical relevance does not resolve situational applicability}
The negative situation changes whether an on-topic preference should be applied. A classifier given the preference and the request marks both sides for use on 99.4\% of pairs. A relevance gate marks both sides relevant on 98.6\% (Table~\ref{tab:gate}). These judgments distinguish topical relevance from situational applicability. An apply head distinguishing relevance from use reaches $\Delta=62.7$.

\section{Discussion}
Recognizing when a preference applies does not ensure that generated answers respect that boundary. Seven models distinguish the situations in multiple choice, while all eight apply the preference on both sides in most free-answer pairs. The dual-track protocol exposes this distinction between selecting and producing an appropriate reply.

Situation removal erases multiple-choice discrimination; memory removal reduces generated uptake. A five-item on-topic pool and two other presentations preserve letter-selection misuse. Stricter prompting changes selection while leaving free-answer always-use at 54\%--85\%. Situational applicability thus remains a distinct target beyond availability and topical relevance.

\paragraph{Possible mechanisms.}
A retrieved preference may be treated as a requirement to satisfy, leading reasoning to focus on how to preserve it before deciding whether it applies. The coffee example illustrates how an answer can accommodate the preference by moving the purchase outside the stated constraint. Longer chains of thought could therefore elaborate such workarounds rather than improve contextual judgment. Unequal attention to preference and situation cues is another possible contributor. Candidate replies may help selection by making the alternative of setting aside the preference explicit. Appendix~\ref{app:mechanisms} develops these hypotheses and proposes tests.

The benchmark studies content preferences supplied in context under English synthetic situations; clean and distractor strata record construction intent. The 1,227-pair collection supports both selection and free generation; reported free-answer rates are conditional on the output-validation and hedging exclusions described in Section~\ref{sec:pending}. Dataset documentation preserves these conditions \citep{gebru2021datasheets,mitchell2019modelcards}.

\section{Conclusion}
PairPref tests when a still-valid preference should guide an answer across 1,227 situation pairs. Its paired design isolates contextual applicability, while two tracks distinguish selection from generation. Models often select appropriate replies yet apply preferences in both generated answers. Evaluating both tracks makes this distinction visible and provides a shared test bed for improving contextual judgment. Reliable personalization requires more than retrieving relevant preferences: it requires judging when they apply and carrying that decision into the answer.

\bibliography{references}
\bibliographystyle{plainnat}
\clearpage
\appendix
\begin{center}
{\large\scshape Technical Appendix}
\end{center}
\vspace{0.8em}
\newcommand{\apptocentry}[3]{%
  \par\noindent\hspace*{#1}\makebox[2.7em][l]{\textbf{\ref*{#2}}}%
  \hyperref[#2]{#3}\nobreak\dotfill\pageref*{#2}\par
}
\begingroup
\setlength{\parskip}{4pt}
\apptocentry{0em}{app:example}{A complete paired example}
\apptocentry{0em}{app:construction_rubric}{Construction checks and model configurations}
\apptocentry{0em}{app:coverage}{Dataset coverage}
\apptocentry{0em}{app:benchmark_comparison}{Comparison with related benchmarks}
\apptocentry{1.5em}{app:benchmark_distinctions}{What the benchmarks measure}
\apptocentry{0em}{app:extended_related}{Extended related work}
\apptocentry{0em}{app:fullreg}{Additional evaluation results and controls}
\apptocentry{1.5em}{app:fullreg-settings}{Implementation details}
\apptocentry{1.5em}{app:fullreg-nomem}{Multiple-choice without memory}
\apptocentry{1.5em}{app:pool-selectivity}{Memory-pool selectivity}
\apptocentry{1.5em}{app:fullreg-far}{Far companions}
\apptocentry{1.5em}{app:fullreg-gate}{Relevance-gate apply head}
\apptocentry{1.5em}{app:fullreg-extra}{Prompt wording}
\apptocentry{1.5em}{app:fullreg-present}{Memory presentation}
\apptocentry{1.5em}{app:cross-track}{Matched comparison across tracks}
\apptocentry{1.5em}{app:coffee-example}{A resource-constrained coffee example}
\apptocentry{1.5em}{app:gen-selectivity}{Free-generation selectivity}
\apptocentry{1.5em}{app:removal-details}{Additional input-removal statistics}
\apptocentry{0em}{app:mechanisms}{Possible mechanisms of preference misuse}
\apptocentry{1.5em}{app:hypothesis-relevance}{Topical relevance as a substitute for applicability}
\apptocentry{1.5em}{app:hypothesis-attention}{Attention to preference and situation cues}
\apptocentry{1.5em}{app:hypothesis-reasoning}{Reasoning directed toward satisfying the preference}
\apptocentry{1.5em}{app:hypothesis-candidates}{Candidate replies as comparison support}
\apptocentry{0em}{app:human_validation}{Human validation of revised materials}
\endgroup
\clearpage

\section{A complete paired example}
\label{app:example}
The following example is drawn from the 1,227-pair evaluation collection.

\paragraph{Target memory.}
User needs food genuinely hot, and finds anything less than mouth-numbing not worth ordering.

\paragraph{Shared request.}
What should I order?

\paragraph{Positive situation.}
I'm choosing my own dinner at the noodle place after clearing the evening, and nobody else is sharing from my bowl.

\paragraph{Negative situation.}
I'm choosing our dinner at the noodle place after clearing the evening, and the person sharing from my bowl has been hiccuping after the table sauce.

\paragraph{Shared replies.}
\begin{description}
\item[$\rmuse$] Order Sichuan hot pot with extra bird's-eye chilies, request the highest heat level, and get chili oil on the side.
\item[$\rmyield$] Order a lighter Sichuan hot pot at medium heat, keep chili oil separate, and let broth carry the flavor in another lane.
\item[$\rmabsent$] Order a mild beef noodle soup with hand-pulled noodles, ask for clear broth, and keep the chili oil on the side.
\item[$\rmoff$] Check whether the restaurant is still delivering tonight. Tony's Pizza usually stops taking online orders at 9:45, even when Google says 10.
\end{description}

\section{Construction checks and model configurations}
\label{app:construction_rubric}
\label{app:rubric}
\label{app:models}
Evaluated data: \path{gen/out/paper_fullreg/verified.jsonl}. The released \texttt{eval\_1227.jsonl} preserves the same preference, request, situations, and candidate replies. Three source collections contribute 161, 549, and 517 uniquely identified pairs.

Generation uses \texttt{gpt-5.5}; verification uses \texttt{deepseek-v4-flash-0731}; the test without context uses \texttt{qwen3.6-flash}. The saved verification fields cover the applicability flip, minimal-pair structure, inference and commonsense requirements, reply roles, acknowledged departure, positive-situation tier, constraint strength, request naturalness and sufficiency, concrete recommendations, category fit, positive-side feasibility, and the availability of answers on both sides. Rule-based filters complement these judgments with checks on text and option form. All retained records have a saved verification verdict of \texttt{PASS}; this records filter acceptance, not independently measured correctness.

The test without context shuffles the four replies and asks the model to select the most appropriate reply from the request and candidates alone. It makes three calls at temperature 1 and retains candidates with at most one selection of the preference-enacting reply. Among retained pairs, 1,056 have zero such selections and 171 have one. These are selection statistics from the construction filter, not a held-out test of robustness to answer cues.

\section{Dataset coverage}
\label{app:coverage}
The collection comprises 1,227 pairs, 45 preferences, eight situation categories, and six domains. Positive situations include 620 clean and 607 distractor cases. These labels describe construction intent.
\begin{table}[h]
\caption{Situation coverage in PairPref.}
\centering\small
\begin{tabularx}{\linewidth}{@{}Xr@{}}
\toprule
Situation category & Pairs \\
\midrule
Companions & 205 \\
Occasion & 163 \\
Time constraint & 136 \\
Beneficiary shift & 296 \\
Role shift & 54 \\
Resource constraint & 110 \\
Physical state & 191 \\
Emotional state & 72 \\
\midrule
Total & 1,227 \\
\bottomrule
\end{tabularx}

\end{table}
Domain counts are food 345, film 236, activities 279, goods 171, routines 104, and social interaction 92. Context-to-request distances of one, three, and six rounds contain 335, 543, and 349 pairs, respectively. A distance of $d$ uses $d-1$ filler exchanges.

\section{Comparison with related benchmarks}
\label{app:benchmark_comparison}
Tables~\ref{tab:benchmark_scope}, \ref{tab:benchmark_protocols_detail}, and~\ref{tab:benchmark_validation} compare evaluation scope and construction evidence. Counts retain each benchmark's unit of evaluation; construction filtering and independent evaluator validation are distinct.
\begin{table}[!ht]
\caption{Protocol details for selected benchmarks in Table~\ref{tab:benchmark_protocols}.}
\label{tab:benchmark_protocols_detail}
\centering
\small
\setlength{\tabcolsep}{4pt}
\renewcommand{\arraystretch}{1.08}
\begin{tabularx}{\linewidth}{@{}>{\raggedright\arraybackslash}p{0.16\linewidth}>{\raggedright\arraybackslash}X>{\raggedright\arraybackslash}X>{\raggedright\arraybackslash}X@{}}
\toprule
Benchmark & Comparison or variation & Boundary assessment & Usefulness assessment \\
\midrule
PrefEval & Explicit or implicit preferences; conversation length & Preference following & Helpfulness in generation \\
RPEval & Single/multiple and explicit/implicit preferences & Intent matching; strategy errors & Feasibility and verbosity errors \\
OP-Bench & Memory methods and memory-free BASE & Irrelevance, sycophancy, repetition & LoCoMo memory utility evaluated alongside \\
BenchPreS & User--recipient--task combinations; with/without preferences & Misapplication and appropriate application & Task completeness, separately scored \\
CIMemories & User attributes across recipient--task contexts & Inappropriate attribute disclosure & Coverage of necessary attributes \\
PersistBench & Leakage, sycophancy, and beneficial-use subsets & Failure rates on targeted scenarios & Separate beneficial-use control subset \\
HUSH-Bench & Matched memory/no-memory; fixed-target invitation pairs & Unsolicited integration, exposure, over-scope & Helpfulness and invited target uptake \\
\midrule
\rowcolor{papergray}
This work & Same memory, request, candidates; context varies & Preference use and withholding & Situational fit and pair success in both tracks \\
\bottomrule
\end{tabularx}
\end{table}

\begin{table}[!ht]
\caption{Scope of related preference and memory benchmarks.}
\label{tab:benchmark_scope}
\centering
\small
\setlength{\tabcolsep}{4pt}
\renewcommand{\arraystretch}{1.08}
\begin{tabularx}{\linewidth}{@{}>{\raggedright\arraybackslash}p{0.22\linewidth}>{\raggedright\arraybackslash}p{0.21\linewidth}>{\raggedright\arraybackslash}X>{\raggedright\arraybackslash}p{0.16\linewidth}@{}}
\toprule
Benchmark & Memory object & Evaluation target & Response \\
\midrule
LoCoMo \citep{maharana2024locomo} & Long dialogues & Recall, event summaries, multimodal continuity & Generation \\
LongMemEval \citep{wu2025longmemeval} & Timestamped histories & Recall, reasoning, updates, abstention & QA \\
PrefEval \citep{zhao2025prefeval} & Explicit or implicit preferences & Preference following over conversations & MCQ; generation \\
PersonaMem \citep{jiang2025personamem} & Evolving user profiles & Current-profile alignment and transfer & MCQ; generation \\
CUPID \citep{kim2025cupid} & Context-dependent preferences & Preference inference and response alignment & Inference; generation \\
\midrule
RPEval \citep{feng2026rpeval} & Preferences linked to query intent & Ignore, support, or dominate & Intent MCQ; generation \\
OP-Bench \citep{hu2026opbench} & Dialogue-derived profiles & Irrelevance, sycophancy, repetition & Generation \\
BenchPreS \citep{yoon2026benchpres} & Response preferences & Selectivity under communication norms & Generation \\
CIMemories \citep{mireshghallah2025cimemories} & Personal information attributes & Context-appropriate disclosure & Generation \\
PersistBench \citep{pulipaka2026persistbench} & User memories and beliefs & Leakage, sycophancy, beneficial use & Generation \\
HUSH-Bench \citep{xu2026hush} & Marked sensitive history & Current-turn warrant for memory use & Generation \\
\midrule
\rowcolor{papergray}
This work & Persistent content preferences & Applicability, behavior, and task completion across contexts & Selection; generation \\
\bottomrule
\end{tabularx}
\end{table}

\begin{table}[!ht]
\caption{Reported scale and validation in the reviewed benchmark papers.}
\label{tab:benchmark_validation}
\centering
\small
\setlength{\tabcolsep}{4pt}
\renewcommand{\arraystretch}{1.18}
\begin{tabularx}{\linewidth}{@{}>{\raggedright\arraybackslash}p{0.17\linewidth}>{\raggedright\arraybackslash}p{0.26\linewidth}>{\raggedright\arraybackslash}X>{\raggedright\arraybackslash}p{0.14\linewidth}@{}}
\toprule
Benchmark & Reported scale and unit & Reported validation & Source locator \\
\midrule
LoCoMo & 50 dialogues in the paper's collection & Human editing for consistency, images, and event grounding & Secs. 1, 3.4; Table 1 \\
LongMemEval & 500 questions; configurable histories & Human rewriting of questions and editing of evidence; judge meta-evaluation & Secs. 3.2--3.3; Table 6 \\
PrefEval & 1,000 base pairs in 3 forms: 3,000 preference--query pairs & Human-assisted curation; manual check of 200 sampled evaluations & Secs. 2.2, 2.5 \\
PersonaMem & About 6,000 query--response pairs; 20 personas & 90 entries from one persona, assessed by three author annotators & Sec. 2; App. B \\
CUPID & 756 instances; 252 per instance type & Human validation and revision; separate preference-matcher meta-evaluation & Secs. 2.3, 3.2--3.3 \\
\midrule
RPEval & 8,255-sample pool; 953-sample finalized test set & Manual verification and adjudication; judge--human ordinal agreement & Secs. 3.1--3.2 \\
OP-Bench & 1,700 instances; 20 users & Two independent reviews per instance; adjudication for disagreement & Secs. 3.3--3.4 \\
BenchPreS & 1,950 attribute-level instances: 10 users, 39 contexts, 5 preferences & Human applicability labels; judge comparison on 100 sampled instances & Secs. 3.2--3.3, 6 \\
CIMemories & 10 evaluated profiles; 49 seed contexts, up to 189 attributes per profile & Persona-conditioned model labels; ambiguous attribute--context pairs excluded & Secs. 4.1, 5.1 \\
PersistBench & 500 samples: 200 leakage, 200 sycophancy, 100 beneficial use & Search-model and held-out-model filtering, then human verification & Secs. 4.1--4.2 \\
HUSH-Bench & 2,400 probe cases; 10 personas & 600 responses, two human annotators each; second-judge evaluation & Secs. 3.2--3.3; Apps. A.1, B.4, D.1 \\
\midrule
\rowcolor{papergray}
This work & 1,227 context pairs; 45 target preferences & Construction filters; two independent annotators assess 200 revised sample pairs & Sec.\ \ref{sec:quality}; App.\ \ref{app:human_validation} \\
\bottomrule
\end{tabularx}
\end{table}

\subsection{What the benchmarks measure}
\label{app:benchmark_distinctions}
PairPref evaluates changes in the applicability of a persistent preference. Its unit is a pair with the same still-valid preference and request, but different situations that warrant applying or setting aside that preference. Selection keeps the candidate replies fixed; free generation removes the candidates. Evaluating both sides measures whether appropriate application and appropriate non-application occur within the same pair. The following comparisons distinguish this target from related evaluations of preference inference, utilization strategies, communication norms, and information disclosure.

\paragraph{CUPID: inferring contextual preferences.}
CUPID constructs interaction histories in which preferences are gradually revealed through user feedback \citep{kim2025cupid}. Context factors, such as people, activities, and tools, are associated with preferences; contrasting factors can have conflicting preferences, and preferences can also change over time. Given a new request and prior sessions, evaluation measures preference inference and response alignment. CUPID therefore already tests contextual personalization. PairPref instead supplies a single preference whose validity is unchanged and tests whether situational facts warrant acting on it. Its contrast concerns whether to apply that preference, rather than which preference the history associates with the current context.

\paragraph{RPEval: choosing a utilization strategy.}
RPEval annotates preference--query instances with \emph{Ignore}, \emph{Support}, or \emph{Dominate} \citep{feng2026rpeval}. Its discriminative task predicts these strategies, while its generative evaluation examines strategy errors and response-level problems such as low feasibility and verbosity. It includes preferences that are topically connected but inappropriate for the task, such as a preference for challenging routes in a family outing with children. Thus, neither contextual applicability nor evaluating discrimination alongside generation is unique to PairPref. PairPref selects among concrete replies held fixed across opposite-applicability situations, and evaluates generated responses on the same pairs, including whether both sides are appropriate.

\paragraph{BenchPreS: expression preferences under communication norms.}
BenchPreS combines 10 user profiles with 39 recipient--task contexts in five formal communication domains \citep{yoon2026benchpres}. Its preference attributes concern role, style, tone, markers, and nickname. Human labels specify which attributes should be applied or suppressed; generated responses are scored using Misapplication Rate and Appropriate Application Rate, with additional task-completeness and prompt-mitigation analyses. It directly measures contextual preference selectivity. PairPref emphasizes preferences governing answer content across eight situation categories and holds the request fixed within each contrast. Its paired scoring checks application and non-application for the same preference--request combination, complementing attribute-level rates across contexts.

\paragraph{OP-Bench: over-personalization failures.}
OP-Bench evaluates irrelevance, sycophancy, and repetition \citep{hu2026opbench}. Its irrelevance subset includes both unrelated queries and deceptively related queries, so its target extends beyond simple topical filtering. Generation is assessed for unnecessary personalization, objectivity, and response diversity; useful memory performance is evaluated alongside on LoCoMo. PairPref instead constructs both an appropriate-use and an inappropriate-use situation for each fixed preference--request pair. This makes selective use within a pair the evaluation unit, rather than combining over-personalization tests with a separate memory-utility benchmark.

\paragraph{CIMemories: contextual information disclosure.}
CIMemories composes personal attributes with recipient--task contexts and labels attributes as necessary or inappropriate to disclose \citep{mireshghallah2025cimemories}. The same attribute can receive different labels across tasks. Evaluation measures inappropriate disclosure and coverage of necessary information, while varying memory composition and examining repeated interactions. This assesses contextual boundaries on information flows. PairPref tests preference-guided content without requiring disclosure.

\paragraph{HUSH-Bench: warrant for sensitive-history use.}
HUSH-Bench tracks marked sensitive disclosures, compares matched memory and no-memory responses, and examines whether retrieved targets enter generated answers \citep{xu2026hush}. Its matched invitation experiment fixes a target-only memory block and adds a generic request to use prior context, measuring target uptake, helpfulness, and over-scope. It therefore already includes controlled paired interventions. PairPref changes situational facts rather than an explicit invitation to use history, and evaluates the appropriateness of applying a persistent preference. Fixed candidate replies additionally support a response-selection test on the same contrasts used for generation.

\paragraph{PersistBench: leakage, sycophancy, and beneficial use.}
PersistBench evaluates cross-domain leakage and memory-induced sycophancy, with a separate beneficial-memory subset that checks whether models can still use helpful information \citep{pulipaka2026persistbench}. Its memory--query cases are constructed to expose failures, and the subsets report their respective failure rates. PairPref links appropriate use and non-use within each fixed preference--request pair. This measures joint success under a situational reversal rather than across separate harmful-use and beneficial-use subsets.

\paragraph{Interpreting the comparison.}
These distinctions concern evaluation objects and protocols, not an absence of contextual reasoning in prior benchmarks. Table~\ref{tab:benchmark_protocols} distinguishes direct capability evaluation from related coverage. Its selection and generation columns concern one still-valid preference and a fixed request, assessed through concrete reply selection or free generation; changing preferences, changing tasks, strategy classification, or explicitly inviting memory use provide related evidence. The outside-scope symbol denotes the evaluation target, not a demonstrated inability of the evaluated models. Disclosure benchmarks remain relevant even when they do not directly evaluate preference enactment. Pair success complements marginal rates: the same per-side accuracy can yield different rates of joint correctness.

\section{Extended related work}
\label{app:extended_related}
\paragraph{Memory organization and control.}
Agent memory research studies how information is formed, retained, and accessed across interactions \citep{hu2025memorysurvey}. Persistent storage and consolidation combine experience retrieval, reflection, and user modeling \citep{park2023generative,zhong2024memorybank}. Memory organization addresses how these records are structured: tiered storage manages information at different scales \citep{packer2023memgpt,kang2025memoryos}, while linked notes and temporal graphs preserve relationships among entries \citep{xu2025amem,rasmussen2025zep}. Retrieval-augmented generation combines language models with external knowledge \citep{lewis2020rag}; HippoRAG and its successor use structured retrieval for knowledge integration \citep{gutierrez2024hipporag,gutierrez2025hipporag2}. Recent approaches address the cost of persistent-memory pipelines \citep{chhikara2025mem0,fang2026lightmem} and adapt memory operations, retrieval actions, or structure selection to the task \citep{yan2026memoryr1,du2025memr3,lu2026fluxmem}. These directions improve the information available to an assistant. We examine the next decision: how an available preference should influence the answer in the current situation.

\paragraph{Memory and personalization benchmarks.}
Existing evaluations test whether an assistant maintains information across interactions and applies it to a user-specific task. Multi-session dialogue and long-term memory benchmarks assess conversational continuity, recall, reasoning, updates, and abstention \citep{xu2022msc,maharana2024locomo,wu2025longmemeval}. Operation-level evaluation distinguishes errors introduced during extraction, updating, and question answering \citep{chen2025halumem}. Personalization benchmarks assess user-conditioned classification and generation, long-form writing, and preference following \citep{salemi2024lamp,kumar2024longlamp,zhao2025prefeval}. Dynamic profiles and contextual preferences extend this setting to changes in user state and interaction history \citep{jiang2025personamem,kim2025cupid}. We hold the preference and request fixed to examine how a change in situation affects the preference's applicability.

\paragraph{Selective personalization and contextual boundaries.}
Appropriate memory use depends on more than topical relevance. Recent benchmarks examine irrational preference integration, over-personalization, and preference suppression under communication norms \citep{feng2026rpeval,hu2026opbench,yoon2026benchpres}. Related work studies cross-domain leakage and memory-induced sycophancy, including mitigation through structured memory presentation \citep{pulipaka2026persistbench,hannoon2026structured}. Sycophancy concerns agreement with user beliefs at the expense of truthfulness \citep{sharma2024sycophancy}; preference applicability concerns whether a recommendation suits the present task. CIMemories evaluates contextual disclosure violations alongside coverage of necessary attributes \citep{mireshghallah2025cimemories}. HUSH-Bench examines unsolicited use of sensitive history, with matched memory/no-memory and explicit-invitation controls \citep{xu2026hush}. These studies establish contextual boundaries as an evaluation target. We test contextual contrasts within a shared preference--request pair and examine what candidate texts reveal about model choices.

\paragraph{Behavioral testing and evaluator reliability.}
Behavioral tests and contrast sets evaluate expected responses to controlled changes \citep{ribeiro2020checklist,gardner2020contrastsets}. A high score need not reflect the intended capability: annotation artifacts and superficial heuristics can provide another route to the answer \citep{mccoy2019hans,gururangan2018artifacts}. Multi-dimensional evaluation makes such distinctions explicit under standardized conditions \citep{liang2023helm}. These principles motivate assessing contextual applicability, candidate cues, and answer completion separately. Language-model judges make this assessment scalable, and the construction checks and response-scoring protocol serve different purposes (Sections~\ref{sec:quality} and~\ref{sec:pending}).

\section{Additional evaluation results and controls}
\label{app:fullreg}

These analyses use the 1,227-pair collection from Section~\ref{sec:legacy}. Each table reports the models with available results for that condition. Use$^+$ and Use$^-$ denote MU$^+$ and MI$^-$, respectively; rates are percentages and $\Delta$ is in percentage points. Generation metrics use the retained pairs defined in Section~\ref{sec:pending}. Dashes indicate unavailable results, not zero scores.

\subsection{Implementation details}
\label{app:fullreg-settings}
The two sides use separate calls. Side-dependent seeds determine memory order, candidate order, and filler exchanges; these presentation differences remain in the no-situation control. Temperature is 0 where supported and omitted at endpoints that reject it. DeepSeek V4 Flash scores every free-answer model, including DeepSeek V4 Pro. Uptake judgments use the preference, request, and answer; separate situational-fit judgments use the current situation, request, and answer.

The controls vary memory availability, situation availability, pool composition, or presentation. No-situation removes situation turns; no-memory removes the memory block. Top-5 retains the target and four near companions; Far-15 retains the target with 15 far companions. Bare removes the relevance clause, while Mitigate explicitly distinguishes a preference's truth from its applicability. User and dialog place the same memories in user messages or prior dialogue. The preference-only classifier receives the target preference and request. The relevance gate also receives the situation and predicts relevance, sufficiency, and applicability separately. Neither receives a generated answer.

\subsection{Multiple-choice without memory}
\label{app:fullreg-nomem}
\begin{table}[!ht]
\caption{Multiple-choice results with the memory block removed.}
\label{tab:nomem-disc}
\centering\small
\begin{tabular*}{\linewidth}{@{\extracolsep{\fill}}lrrr@{}}
\toprule
Model & MU$^+$ & MI$^-$ & $\Delta$ \\
\midrule
Gemini 3.5 Flash & 39.7 & 2.4 & 37.3 \\
GPT-6 Astra & 32.5 & 2.2 & 30.3 \\
Kimi K3 & 34.1 & 2.9 & 31.1 \\
GLM-5.3 & 34.5 & 2.2 & 32.3 \\
DeepSeek V4 Pro & 30.2 & 2.5 & 27.7 \\
\bottomrule
\end{tabular*}
\end{table}

Figure~\ref{fig:controls}a shows the same comparison. Without a memory block, negative-side selection of $\rmuse$ is 2.2\%--2.9\%. Positive-side selection remains 30.2\%--39.7\%. The remaining gap shows that situation and candidate content can favor the preference-enacting reply even without an explicit memory block; it does not demonstrate retrieval of a withheld preference.

\subsection{Memory-pool selectivity}
\label{app:pool-selectivity}
\begin{table}[!ht]
\caption{Memory-pool controls: negative-side use and selectivity.}
\label{tab:pool-selectivity}
\centering\small
\begin{tabular*}{\linewidth}{@{\extracolsep{\fill}}lrrrr@{}}
\toprule
& \multicolumn{2}{c}{Use$^-$} & \multicolumn{2}{c}{$\Delta$} \\
\cmidrule(lr){2-3}\cmidrule(lr){4-5}
Model & Top-16 & Top-5 & Top-16 & Top-5 \\
\midrule
Gemini 3.5 Flash & 22.9 & 25.8 & 56.0 & 55.7 \\
DeepSeek V4 Pro & 23.0 & 27.8 & 53.5 & 50.8 \\
Kimi K3 & 18.3 & 20.7 & 57.6 & 59.0 \\
GLM-5.3 & 18.4 & 21.0 & 56.6 & 56.6 \\
GPT-6 Astra & 14.7 & 11.7 & 56.6 & 61.5 \\
\bottomrule
\end{tabular*}
\end{table}

Table~\ref{tab:pool-selectivity} supplements Table~\ref{tab:pool} with selectivity ($\Delta$) for the same pool-size controls.

\subsection{Far companions}
\label{app:fullreg-far}
\begin{table}[!ht]
\caption{Multiple-choice results with the target and 15 far companions.}
\label{tab:far15}
\centering\small
\begin{tabular*}{\linewidth}{@{\extracolsep{\fill}}lrrrr@{}}
\toprule
& \multicolumn{2}{c}{MI$^-$} & \multicolumn{2}{c}{$\Delta$} \\
\cmidrule(lr){2-3}\cmidrule(lr){4-5}
Model & Far-15 & Top-16 & Far-15 & Top-16 \\
\midrule
Gemini 3.5 Flash & 29.2 & 22.9 & 53.5 & 56.0 \\
DeepSeek V4 Pro & 23.6 & 23.0 & 52.8 & 53.5 \\
Kimi K3 & 22.9 & 18.3 & 56.4 & 57.6 \\
GLM-5.3 & 21.1 & 18.4 & 55.5 & 56.6 \\
GPT-6 Astra & 12.8 & 14.7 & 62.4 & 56.6 \\
\bottomrule
\end{tabular*}
\end{table}

Relative to Top-16, Far-15 changes $\Delta$ by $-2.5$ to $+5.8$ points. Negative-side use does not collapse when the extra memories are unrelated. Both pool edits preserve the main selection pattern (Table~\ref{tab:pool}).

\subsection{Relevance-gate apply head}
\label{app:fullreg-gate}
\begin{table}[!ht]
\caption{Answer-free classifiers on the same 1,227 pairs.}
\label{tab:gate}
\centering\small
\begin{tabular*}{\linewidth}{@{\extracolsep{\fill}}lrrr@{}}
\toprule
Condition & MU$^+$ & MI$^-$ & $\Delta$ \\
\midrule
Preference-only & 99.4 & 99.4 & 0.0 \\
Gate apply & 87.9 & 25.2 & 62.7 \\
\midrule
Both sides relevant & \multicolumn{3}{c}{98.6} \\
Both sides sufficient & \multicolumn{3}{c}{61.3} \\
\bottomrule
\end{tabular*}
\end{table}

The preference-only classifier makes one decision per preference--request pair without seeing either situation; its 99.4\% application rate therefore appears on both sides. The relevance gate sees each situation separately. Both-sides relevance and sufficiency require the corresponding decision to be positive on both sides. The apply output explicitly distinguishes relevance from applicability; its $\Delta$ measures contextual application decisions, not retrieval performance.

\subsection{Prompt wording}
\label{app:fullreg-extra}
\begin{table}[!ht]
\caption{Effect of prompt wording on multiple-choice and free-answer performance.}
\label{tab:prompt}
\centering\small
\begin{tabular*}{\linewidth}{@{\extracolsep{\fill}}lrrrrrr@{}}
\toprule
& \multicolumn{3}{c}{MCQ MI$^-$} & \multicolumn{2}{c}{Always-use} & \\
\cmidrule(lr){2-4}\cmidrule(lr){5-6}
Model & Nudge & Bare & Mitigate & Nudge & Mitigate & \shortstack{\scriptsize Gen-OK\\\scriptsize mitigate} \\
\midrule
Claude Opus 5 & 11.5 & 11.5 & 3.6 & 88.7 & --- & --- \\
GPT-6 Astra & 14.7 & 16.3 & 7.9 & 71.0 & 54.0 & 26.8 \\
Kimi K3 & 18.3 & 20.0 & 12.0 & 83.6 & 85.5 & 9.5 \\
GLM-5.3 & 18.4 & 20.0 & 8.1 & 90.8 & 85.2 & 10.3 \\
Qwen 3.8 Max & 20.3 & 22.2 & 11.0 & 81.2 & --- & --- \\
Gemini 3.5 Flash & 22.9 & 25.3 & 19.0 & 75.1 & 72.5 & 20.8 \\
DeepSeek V4 Pro & 23.0 & 25.7 & 15.2 & 81.3 & 84.6 & 11.9 \\
Grok 4.6 & 57.4 & 57.0 & 39.0 & 92.0 & --- & --- \\
\bottomrule
\end{tabular*}
\end{table}

\begin{table}[!ht]
\caption{Positive-side multiple-choice use under nudge versus mitigate.}
\label{tab:prompt-mu}
\centering\small
\begin{tabular*}{\linewidth}{@{\extracolsep{\fill}}lrr@{}}
\toprule
Model & Nudge & Mitigate \\
\midrule
Claude Opus 5 & 76.4 & 67.6 \\
GPT-6 Astra & 71.2 & 66.0 \\
Kimi K3 & 76.0 & 68.9 \\
GLM-5.3 & 75.1 & 66.1 \\
Qwen 3.8 Max & 71.6 & 67.6 \\
Gemini 3.5 Flash & 78.9 & 80.3 \\
DeepSeek V4 Pro & 76.4 & 72.4 \\
Grok 4.6 & 75.1 & 69.6 \\
\bottomrule
\end{tabular*}
\end{table}

Figure~\ref{fig:prompt} summarizes these comparisons. The mitigate instruction reduces positive-side multiple-choice use for seven of eight models. Flash is the exception (78.9\% to 80.3\%). Most models thus reduce both appropriate and inappropriate use, creating a trade-off.

\subsection{Memory presentation}
\label{app:fullreg-present}
\begin{table}[!ht]
\caption{Effect of memory presentation on negative-side multiple-choice use.}
\label{tab:present}
\centering\small
\begin{tabular*}{\linewidth}{@{\extracolsep{\fill}}lrrr@{}}
\toprule
Model & System list & User block & Dialog turns \\
\midrule
Claude Opus 5 & 11.5 & 12.0 & 12.7 \\
GPT-6 Astra & 14.7 & 16.3 & 20.1 \\
Kimi K3 & 18.3 & 19.6 & 20.6 \\
GLM-5.3 & 18.4 & 20.0 & 20.9 \\
Qwen 3.8 Max & 20.3 & 23.1 & 27.1 \\
Gemini 3.5 Flash & 22.9 & 24.6 & 25.6 \\
DeepSeek V4 Pro & 23.0 & 26.2 & 27.5 \\
Grok 4.6 & 57.4 & 58.4 & 58.4 \\
\bottomrule
\end{tabular*}
\end{table}

Figure~\ref{fig:controls}b reports these rates; dialog raises Qwen 3.8 Max from 20.3\% to 27.1\%.

\subsection{Matched comparison across tracks}
\label{app:cross-track}
\label{app:track-examples}
We join default-condition selection and generation records by sample identifier within each model. Common pairs have valid uptake judgments on both generated answers and neither answer flagged as hedging. Selection success requires $\rmuse$ on the positive side and $\rmyield$ or $\rmabsent$ on the negative side. The final column divides pairs with $u^+\ge0.5$ and $u^-\ge0.5$ by selection-successful common pairs. Situational-fit judgments are not required for this uptake comparison.
\begin{table}[ht]
\caption{Preference use in generation after successful paired selection.}
\centering\small
\begin{tabular*}{\linewidth}{@{\extracolsep{\fill}}lrrrr@{}}
\toprule
Model & Common pairs & Selection success & Use both & Rate (\%) \\
\midrule
Claude Opus 5 & 1067 & 719 & 641 & 89.2 \\
Kimi K3 & 1091 & 654 & 561 & 85.8 \\
GPT-6 Astra & 1068 & 622 & 455 & 73.2 \\
GLM-5.3 & 1056 & 615 & 567 & 92.2 \\
Gemini 3.5 Flash & 1065 & 613 & 440 & 71.8 \\
DeepSeek V4 Pro & 1107 & 624 & 517 & 82.9 \\
Qwen 3.8 Max & 1011 & 549 & 446 & 81.2 \\
Grok 4.6 & 1153 & 272 & 245 & 90.1 \\
\bottomrule
\end{tabular*}
\end{table}

\subsection{A resource-constrained coffee example}
\label{app:coffee-example}
The example is \texttt{M07-C6-c26634}, evaluated with Claude Opus 5. It illustrates a scored failure and is not used to estimate its frequency.

\paragraph{Preference.}
User only drinks hand-poured single-origin, and rules out instant and chain coffee.

\paragraph{Shared request.}
What should I order?

\paragraph{Positive situation.}
I'm settled at the market hall counter and they asked me to pick, with the afternoon clear and my wristband still loaded for whatever they can make.

\paragraph{Negative situation.}
I'm settled at the market hall counter and they asked me to pick, with the afternoon clear and my wristband showing only enough credit for the self-serve urn.

\paragraph{Selected positive reply.}
Order the Ethiopian pour-over, preferably a washed Yirgacheffe if they have it. Ask for it black so the florals stay clear.

\paragraph{Selected negative reply.}
Order the house drip coffee, medium roast if labeled. Take it black first, then add a small splash if it tastes harsh.

\paragraph{Generated positive answer.}
Ask whoever's on bar which single-origin they've got set up for pour-over right now, and let them pick the one they're most excited about. That usually lands better than reading the board, and it's exactly the kind of thing a counter person enjoys being asked. Worth asking how it's prepped too — whether it's a V60, a Kalita, whatever they're dialed in on that day.

For food, see if there's a regional pastry from a small baker in the hall rather than the standard case stuff. And since you're in a market hall with a clear afternoon, the fermented and preserved stalls are worth a slow walk after the coffee — pickles, krauts, anything cured. That's a good browse even if you only end up with one jar.

Keeps the whole thing modest, and you'll leave knowing something you didn't walk in with.

\paragraph{Generated negative answer.}
Skip the urn. It's the one thing on offer that doesn't fit how you drink coffee — batch-held drip isn't going to resemble a hand-poured single origin, and paying for it just to have a cup in hand is a poor trade.

What the wristband is actually good for at a market hall counter is food. Ask whoever's behind it what's made in-house and what came in from a producer they know — that's usually where the interesting answer lives. A regional pastry or something from the ferment/preserve side of the counter tends to be the better use of limited credit than a mediocre coffee, and it's the kind of thing you can ask a real question about: who bakes it, what the starter or brine is, whether it's a local tradition.

Then let the coffee wait. Your afternoon is clear and three kilometres is only 35–40 minutes on foot, so there's easy room to walk to a place that pours to order. Worth asking the counter staff directly where they go on their own break — that'll get you somewhere better than searching ratings.

The saved scores are $u^+=u^-=1$, with situational fit 1 on the positive side and 0 on the negative side; neither generated answer is flagged as hedging.

\subsection{Free-generation selectivity}
\label{app:gen-selectivity}
Table~\ref{tab:gen-selectivity} reports the generation uptake gap omitted from the main table. These values use the generation scoring rule and the same retained pairs as the main results.
\begin{table}[ht]
\caption{Selectivity in free generation.}
\label{tab:gen-selectivity}
\centering\small
\begin{tabular}{lr}
\toprule
Model & $\Delta$ (points) \\
\midrule
Claude Opus 5 & 1.5 \\
Kimi K3 & 6.8 \\
GPT-6 Astra & 10.9 \\
GLM-5.3 & 3.3 \\
Gemini 3.5 Flash & 14.0 \\
DeepSeek V4 Pro & 10.7 \\
Qwen 3.8 Max & 8.3 \\
Grok 4.6 & 2.8 \\
\bottomrule
\end{tabular}
\end{table}

\subsection{Additional input-removal statistics}
\label{app:removal-details}
\begin{table}[!ht]
\caption{Additional statistics for input-removal controls.}
\label{tab:validity-details}
\centering\small
\begin{tabular*}{\linewidth}{@{\extracolsep{\fill}}lrrrrrrr@{}}
\toprule
& \multicolumn{4}{c}{No situation} & \multicolumn{3}{c}{No memory, free answer} \\
\cmidrule(lr){2-5}\cmidrule(lr){6-8}
Model & MCQ $\Delta$ & Gen $\Delta$ & MCQ Use$^+$ & MCQ Use$^-$ & $\Delta$ & MU$^+$ & MI$^-$ \\
\midrule
Gemini 3.5 Flash & 0.0 & $-1.8$ & 90.0 & 90.0 & 3.6 & 11.1 & 7.5 \\
Kimi K3 & $-0.2$ & $-1.0$ & 89.6 & 89.9 & 1.9 & 15.1 & 13.1 \\
DeepSeek V4 Pro & $-0.5$ & 0.4 & 89.1 & 89.6 & 2.4 & 11.7 & 9.3 \\
GLM-5.3 & $-0.7$ & $-0.3$ & 87.2 & 87.9 & 0.5 & 16.3 & 15.8 \\
GPT-6 Astra & 1.4 & 2.1 & 83.9 & 82.6 & 0.1 & 11.4 & 11.3 \\
Qwen 3.8 Max & --- & $-0.4$ & --- & --- & 0.3 & 11.7 & 11.4 \\
\bottomrule
\end{tabular*}
\end{table}

The no-situation Use columns concern selection. Dashes denote unreported results, not zero scores. Generation applies paired output-validation and hedging exclusions per condition.

\section{Possible mechanisms of preference misuse}
\label{app:mechanisms}
PairPref reveals a gap between selecting an appropriate reply and producing one under the same situations. Several mechanisms could contribute to this pattern, and they may operate together. The following hypotheses connect the behavioral findings to predictions that future experiments can test; they are not measurements of the evaluated models' internal computation.

\subsection{Topical relevance as a substitute for applicability}
\label{app:hypothesis-relevance}
A model may treat a true preference that matches the request as sufficient grounds for using it. In PairPref, both properties remain intact even when the situation makes preference use inappropriate. The situation-blind classifier recommends application for 99.4\% of pairs, while the gate marks both sides relevant for 98.6\%. These results illustrate why relevance alone cannot resolve applicability. One hypothesis is that generation relies on a similar shortcut. A direct test would compare relevance-only instructions with a separate applicability decision conditioned on the full situation, followed by answer generation. Scoring the intermediate decision and the final answer separately would distinguish incorrect judgments from failures to follow a correct judgment. Evaluation should measure appropriate use and misuse with inputs fixed.

\subsection{Attention to preference and situation cues}
\label{app:hypothesis-attention}
The retrieved preference may exert a more persistent influence during decoding than the situational constraint. A preference directly specifies desired answer content, whereas recognizing a constraint can require connecting several details. Unequal attention to these cues is therefore a plausible contributor. Misuse with shorter pools and alternative presentations suggests that a simple account based only on context length or one memory location is insufficient. Follow-up experiments could vary cue position and explicitness while preserving meaning. For models with accessible internals, attention and activation analyses could identify candidate components, followed by targeted interventions to test whether they change contextual preference use. Attention weights alone cannot establish causality; interventions should reduce misuse while preserving appropriate use.

\subsection{Reasoning directed toward satisfying the preference}
\label{app:hypothesis-reasoning}
Once a model adopts preference satisfaction as its objective, additional reasoning may produce ways to preserve the preference despite the situation. The coffee example in Appendix~\ref{app:coffee-example} is consistent with this possibility: the answer proposes another shop or a later purchase rather than selecting within the stated budget. This behavior motivates a distinction between reasoning about how to satisfy a preference and reasoning about whether to apply it. A controlled comparison could evaluate direct answering, generic step-by-step reasoning, and an explicit applicability-first procedure under comparable generation budgets. The hypothesis predicts that generic reasoning need not improve pair success, whereas checking applicability first may reduce workarounds. Written rationales may reveal justification patterns without establishing the underlying computation.

\subsection{Candidate replies as comparison support}
\label{app:hypothesis-candidates}
Selection supplies explicit alternatives, including replies that set the preference aside while meeting the request. Free generation requires the model to construct those alternatives itself. This difference may help explain why a model selects the intended replies yet uses the preference in both generated answers for the same pair. A follow-up could compare direct generation with a procedure that first drafts both a preference-enacting answer and a preference-withholding answer, then chooses between them using the situation. Comparing model-generated alternatives with the benchmark candidates would help separate difficulty producing an appropriate alternative from difficulty choosing it. Evaluation should retain the generation track's situational-fit criteria: withholding a preference alone does not ensure an appropriate answer.

\section{Human validation of revised situations and candidate replies}
\label{app:human_validation}

\paragraph{Sample and protocol.}
Two annotators independently assessed 200 pairs spanning 45 preferences, eight situation categories, and six domains. Sampling assigned one pair per preference, then allocated 155 proportionally by largest remainder and sampled without replacement. Before annotation, 112 situations in 107 pairs were revised; preferences, requests, and replies were unchanged. This assessment concerns revised materials, separately from the 1,227-pair model evaluation.

Sheets randomized situations and options and hid original labels and automated judgments. Annotators first judged 400 situations for applicability, continued preference validity, and request naturalness and answerability, then checked pair control and selected all acceptable replies. Labels allowed uncertainty, distinguished from an empty acceptable set. Results retain independent, pre-adjudication judgments.

\paragraph{Inter-annotator agreement.}
Table~\ref{tab:human_agreement} reports exact agreement and unweighted Cohen's $\kappa$. Applicability agreement is 99.25\%, with $\kappa=0.985$. Resampling complete pairs 5,000 times gives descriptive percentile 95\% intervals of [98.25\%, 100.00\%] for agreement and [0.965, 1.000] for $\kappa$, preserving dependence between the two situations in a pair.

\begin{table}[htbp]
\centering\small
\caption{Inter-annotator agreement on revised materials.}
\label{tab:human_agreement}
\begin{tabularx}{\textwidth}{@{}Xrrr@{}}
\toprule
Dimension & Judgments & Agreement (\%) & Cohen's $\kappa$ \\
\midrule
Preference applicability & 400 & 99.25 & 0.985 \\
Continued preference validity & 400 & 100.00 & --- \\
Request naturalness and answerability & 400 & 97.75 & $-0.010$ \\
Controlled pair comparison & 200 & 99.50 & 0.000 \\
Candidate acceptability & 1,580 & 87.72 & 0.728 \\
\bottomrule
\end{tabularx}
\par\vspace{3pt}
\begin{minipage}{\textwidth}\footnotesize
Candidate judgments exclude five situations with an uncertain label from either annotator. A dash denotes undefined $\kappa$ because both annotators used a single category.
\end{minipage}
\end{table}

Both annotators marked all 400 preferences as still valid. For request answerability, they marked 397 and 394 situations as acceptable, respectively; their three and six negative judgments did not overlap. The observed agreement of 97.75\% is slightly below the 97.77\% expected from these marginal distributions, yielding $\kappa=-0.010$. For pair control, one annotator accepted all 200 pairs and the other accepted 199, yielding $\kappa=0$. These distributions explain why high raw agreement can coexist with low or undefined $\kappa$: agreement on the dominant category does not establish agreement on the exceptions.

For candidate acceptability, each selected option is treated as accepted and each unselected option as rejected. The five uncertain situations are excluded, leaving 395 situations and 1,580 binary option judgments. An explicitly empty set contributes four rejections. Exact agreement on the complete acceptable set is 62.0\% across all 400 situations. Mean Jaccard similarity is 0.771 across the 395 jointly determinate situations, with two empty sets assigned similarity one.

\begin{table}[htbp]
\centering\small
\caption{Joint human confirmation of labels and candidate replies.}
\label{tab:human_support}
\begin{tabularx}{\textwidth}{@{}Xrr@{}}
\toprule
Criterion & Count & Rate (\%) \\
\midrule
\multicolumn{3}{@{}l}{\textit{Situations and labels}} \\
Original positive side is applicable & 183/200 & 91.5 \\
Original negative side is inapplicable & 180/200 & 90.0 \\
Both sides match the original direction & 180/200 & 90.0 \\
Applicability contrast, either direction & 196/200 & 98.0 \\
\midrule
\multicolumn{3}{@{}l}{\textit{Candidate replies}} \\
Positive-side $\rmuse$ is acceptable & 157/200 & 78.5 \\
Negative-side $\rmuse$ is unacceptable & 165/200 & 82.5 \\
No candidate is acceptable & 20/400 & 5.0 \\
\bottomrule
\end{tabularx}
\par\vspace{3pt}
\begin{minipage}{\textwidth}\footnotesize
Each row requires confirmation from both annotators. Either-direction contrasts include 16 pairs opposite to the original labels. Rates are unweighted sample statistics.
\end{minipage}
\end{table}

\paragraph{Support for the original labels.}
Table~\ref{tab:human_support} retains the original positive/negative mapping, verified against identifiers and both annotation sheets. Both annotators confirm the original direction for 180 pairs and the opposite direction for 16 revised pairs. Thus, 98.0\% confirm a contrast, while 90.0\% support its original direction. All 93 unchanged pairs receive joint support for their original direction.

\paragraph{Candidate quality and scope.}
Candidate judgments are less uniform than applicability judgments. Both annotators find no acceptable reply in 20 situations; at least one does so in 46. These cases and the direction reversals require review, not automatic relabeling. The reported rates describe the revised sample, not correctness of the original 1,227 pairs. Incorporating revisions requires updating affected model evaluations. No generated answers were annotated, so these results do not validate the automatic free-generation scorer.

\end{document}